\documentclass[a4paper]{article}

\usepackage{INTERSPEECH2022}

\usepackage{amsmath,graphicx}
\usepackage{verbatim}
\usepackage{booktabs} 
\usepackage{caption}
\usepackage{xcolor}
\usepackage{pifont}
\usepackage{subcaption}
\usepackage{enumitem}
\usepackage{hyperref}
\usepackage{svg}
\usepackage{pmboxdraw}
\usepackage{nicefrac}

\title{Device Directedness with Contextual Cues for Spoken Dialog Systems}
\name{Dhanush Bekal, Sundararajan Srinivasan, Sravan Bodapati, Srikanth Ronanki, Katrin Kirchhoff}
\address{
  AWS AI Labs}
\email{\{dkannang, sundarsr, sravanb, ronanks, katrinki\}@amazon.com}

\begin{document}

\maketitle
\begin{abstract}

In this work, we define barge-in verification as a supervised learning task where audio-only information is used to classify user spoken dialogue into true and false barge-ins. Following the success of pre-trained models, we use low-level speech representations from a self-supervised representation learning model for our downstream classification task.
Further, we propose a novel technique to infuse lexical information directly into speech representations to improve the domain-specific language information implicitly learned during pre-training. Experiments conducted on spoken dialog data show that our proposed model trained to validate barge-in entirely from speech representations is faster by 38\% relative and achieves 4.5\% relative F1 score improvement over a baseline LSTM model that uses both audio and Automatic Speech Recognition (ASR) 1-best hypotheses. On top of this, our best proposed model with lexically infused representations along with contextual features provides a further relative improvement of 5.7\% in the F1 score but only 22\% faster than the baseline.
\end{abstract}
\noindent\textbf{Index Terms}: Spoken dialog systems, barge-in, speech representations.


\section{Introduction}
\label{sec:introduction}




Voice based chat-bots have become a ubiquitous part of commercial goal oriented dialogue systems \cite{pieraccini2009we, budzianowski2018multiwoz, hosseini2020simple}. 
These dialog systems in general have an additional barge-in feature \cite{strom2000intelligent} which allows customers to interrupt the ongoing dialogue at any point of time to reach the goal as quickly as possible. However, barge-in systems \cite{strom2000intelligent, rose2003hybrid} are quite sensitive to background speech and non chat-bot directed customer speech (e.g., whispering or talking to others) resulting in a high number of false positives. Reducing these false positives helps improve customer experience by avoiding unnecessary processes in the downstream dialog system. In our work, we define true barge-in as those speech utterances with interruptions directed at the chat-bot and any other detected speech is classified as a false barge-in (see Table \ref{tab:examples}). This is akin to device-directed speech detection applied to close talk chat-bot dialogue systems with an added task of barge-in (or interruption) detection. However, our task is different from device directed speech detection in the literature \cite{shriberg2012learning, mallidi2018device, norouzian2019exploring} as we do not deal with far field audio or wake-word recognition. 


Traditionally, barge-in is handled in three stages: detection, verification, and recovery \cite{strom2000intelligent}. Detection is the identification of speech input from the user and the problems here include accurately detecting foreground voice activity in the presence of background speech, noise, music, and other audio events. Verification means determining whether the utterance is actually meant to be a barge-in, or something else (e.g., non chat-bot directed speech) and generally requires information about the lexical content of the utterance. Some systems \cite{selfridge2013continuously} use confidence scores from incremental ASR results (partial decoding hypotheses) for this purpose, others also make use of NLU results, or use special language model arcs. It has been shown that model-based verification improves barge-in detection over pure voice activity detection (VAD) based approaches \cite{rose2003hybrid}. Including user-specific information such as typical barge-in rates and average ASR accuracy for a given user have also been shown to help during the verification stage \cite{komatani2010online}.


\begin{table}[t]
\caption{Examples of True and False Barge-in in a dialogue}
  \vspace{-3mm}
  \centering
  \small
    \begin{tabular}{l|p{5cm}}
        \toprule
        \textbf{No barge-In} & \textbf{Bot} : Would you like to book a car ? \\
        & \textbf{User} : Yes \\
        & \textbf{Bot} : Would you like a Sedan, SUV, Hatchback ? \\
        & \textbf{User} : Sedan \\
        \midrule
        \textbf{True Barge-In} & \textbf{Bot} : Would you like to book... \\
        & \textbf{User} : Yes \\
        & \textbf{Bot} : Would you like a Sedan... \\
        & \textbf{User} : Sedan \\
        \midrule
        \textbf{False Barge-In} & \textbf{Bot} : Would you like a Sedan... \\
        & \textbf{User} : Can you get me a coffee? \\
     \bottomrule
    \end{tabular}
    \label{tab:examples}
\vspace{-5mm}
\end{table}

One potential drawback of ASR/NLU model-driven verification is latency and in general, human-computer interaction (HCI) is significantly impacted by delayed responses from a cascaded spoken dialogue system. Also, NLU models often employ data-efficient techniques to make them robust to ASR errors. Hence, end-to-end (E2E) spoken language understanding (SLU) solutions have recently been proposed to address cascading errors as well as to decrease latency \cite{tian2020improving, cao2021sequential, saxon2021end}. Similarly, our work focuses on using audio and other dialogue information available prior to an ASR system for barge-in verification. Although E2E solutions are elegant and straightforward, they are often not considered for practical use due to a gap in performance over a cascaded system. In this work, we show that self-supervised representation learning models like Hidden-Unit BERT (HuBERT) \cite{hsu2021hubert} that implicitly encode linguistic information can be utilized for the barge-in classification task without compromising on performance. Experimental results show that our best proposed model to validate barge-ins directly from the audio is not only significantly better (10.4\% relative F1 improvement: 73.6 to 81.3) but also 22\% faster than a baseline system that uses both audio and ASR hypothesis. The main contributions of our work are summarized as follows:
\begin{itemize}[leftmargin=*,itemsep=1pt, topsep=2pt]
    \item Pre-trained speech representations derived directly from the audio can be utilized for barge-in classification with improved accuracy and latency over a baseline model that uses lexical information from ASR hypothesis. 
    \item We introduce a novel technique to infuse lexical information directly into HuBERT-based speech representations and show that it helps improve the performance of our downstream end-to-end SLU classification task.
    \item We show that contextual features such as bot prompt and dialogue context play a prominent role in identifying true barge-in interruptions. 
\end{itemize}

\begin{figure} 
  \centering
  \includegraphics[scale=0.75]{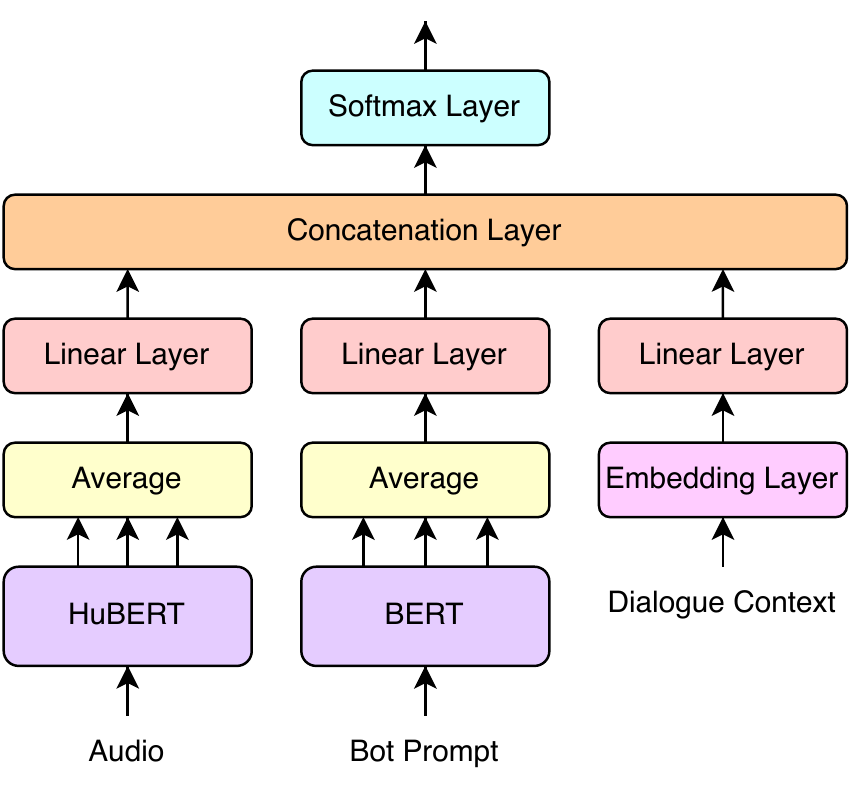}
  \vspace{-3mm}
  \caption{Schematic diagram of contextual barge-in model}
  \label{fig:model1}
  \vspace{-5mm}
\end{figure}

\section{Proposed Approach}
\label{sec:proposed_approach}

Figure \ref{fig:model1} illustrates the contextual model we use for the barge-in classification task. The model encodes three types of input features: (i) audio encoder using a pretrained HuBERT model, (ii) bot prompt encoder using a pretrained BERT model, and (iii) dialogue context encoder using a trainable embedding layer. All three encoded representations are joined by a concatenation layer and then passed through a softmax layer for classification.


\subsection{Audio Encoder for Speech Representation}
\label{ssec:audio_encoder}

Self-supervised learning (SSL) has shown prominent results for speech processing, especially on phoneme classification and ASR \cite{van2018representation, baevski2020wav2vec, ling2020decoar}. 
Recently, \cite{yang2021superb} proposed the SUPERB benchmark \footnote{\url{https://superbbenchmark.org/leaderboard}} to evaluate SSL models across different tasks and as per the results, HuBERT \cite{hsu2021hubert} and WavLM \cite{chen2021wavlm} (both use masked prediction loss) enjoy the best generalization ability in the overall evaluation. HuBERT utilizes an offline clustering step to generate noisy target labels for a BERT-like pre-training \cite{devlin2018bert}. The loss function used for HuBERT pre-training is a combination of unmasked and masked prediction loss. The cross entropy loss computed only over unmasked time steps is similar to acoustic modeling and the loss computed only over masked time steps where the model has to predict the targets corresponding to unseen frames from context, is analogous to language modeling. This way the HuBERT model learns both the acoustic representation of unmasked segments and the long-range temporal structure of the speech data \cite{srinivasan2021representation}. Prior literature shows that linguistic information helps in tasks like device directed speech detection, and therefore we leverage these implicit language representations for our barge-in classification task. 

Let $X$ denote a speech utterance $X = \{x_1, x_2,...x_T\}$ of T frames. The HuBERT-based audio encoder transforms the speech sequence $X$ into a sequence of hidden representations
$H_x=\{h_{x_1},h_{x_2},...h_{x_M}\}$. We then compute the mean of the ensemble and pass it through a linear layer to obtain a single speech representation $r_x$, which is defined as:

\begin{equation}
\label{eq:speech_rep}
    r_x = W^x (\frac{1}{T}\sum_{i=1}^{T}h_{x_i})
\end{equation}

\noindent where $W^x$ denotes the weight matrix of the linear layer. For audio-only experiments presented in section \ref{sec:results}, this speech representation is directly passed through a softmax layer. 

\subsection{Bot Prompt and Dialog Context Encoder}
\label{ssec:context_encoder}

Spoken dialog systems typically span multiple turns of back and forth between a user and a bot. These interactions mostly adhere to a dialogue structure that can be determined beforehand. \textit{Bot prompt} and \textit{dialogue context} are two such contextual dialogue inputs available prior to processing user utterance for barge-in. 



\vspace{2mm}
\noindent\textbf{Bot Prompt Representation:} Bot prompts are the questions asked by a chat-bot to user in spoken dialog systems. Examples of a single turn bot prompt and user response are given in Table \ref{tab:examples}. The intuition behind using bot prompt information is that they are correlated to the user response in true barge-in cases i.e., users interrupt with responses that are expected by the chat-bot. For example, when the chat-bot is asking a question, there is a higher chance of interruption in certain dialogue states if the user has knowledge of the interaction. In order to extract prompt representations, we tokenize bot prompts into sub-words and encode them using a pretrained BERT model \cite{devlin2018bert}. 

Let $P$ denote a bot prompt utterance $P = \{p_1, p_2,...p_N\}$ of N tokens. The BERT-based bot prompt encoder transforms the prompt sequence $P$ into a sequence of hidden representations $H_p=\{h_{p_1},h_{p_2},...h_{p_N}\}$. The ensembeled average is then passed through a linear layer whose weights are defined by $W^p$ to obtain a single prompt representation $r_p$, which is defined as:
\begin{equation}
    r_p = W^p (\frac{1}{N}\sum_{i=1}^{N}h_{p_i})
\end{equation}

\noindent\textbf{Dialogue Context Representation:} We create dialogue context labels associated with each bot prompt. These dialogue context labels indicate whether the chat-bot expects an intent, and a slot in the user response to the bot prompt. We use 3 intents and 7 slots resulting in a total of 10 unique dialogue context labels. To represent these dialogue contexts ($r_d$), we embed them using a trainable embedding layer followed by a linear projection layer and is defined as: 

\begin{equation}
    r_d = W^d \bar{E}D
\end{equation}

\noindent where $\bar{E} \in \mathbb{R}^{d \mathrm{x} m}$ is the dialogue context embedding matrix. $m$ and $d$ are the dialogue embedding dimensionality and the number of unique dialog context labels respectively. 


\subsection{Combined Model}
As shown in figure \ref{fig:model1}, all three representations are concatenated and then passed through a feed forward layer with \textit{tanh} activation followed by a softmax layer to get the final output ($\hat{y}$): 

\begin{equation}
    r_c = tanh(W^c(r_x \oplus r_p \oplus r_d) + b^c) 
\end{equation}
\begin{equation}
    \hat{y} = softmax(W^yr_c + b^y) 
\end{equation}

\noindent where $W^c$, $W^y$, $b^c$, $b^y$ denote weight matrices and bias of the concatenation and classification layer respectively. The model is trained using a cross entropy loss function. 

\section{Language Infusion into Speech Representation}
For improving the semantic information implicitly learned during HuBERT pre-training, we propose a novel approach to infuse lexical information directly into speech representations. Figure 2 shows the overview of our language infusion approach. For each training utterance, the corresponding time-aligned word transcript is obtained either using forced alignment of human labels or using an ASR system that returns words with time-stamps. The words are tokenized to sub-words and run through a pre-trained BERT model. Each word $w$ is then represented by the mean of the BERT embeddings corresponding to the sub-word tokens belonging to the word. On the acoustic side, the speech representation is first extracted from a pre-trained HuBERT-base model. The frame-level features are then passed through a multi-layer transformer \cite{vaswani2017attention} (this is the core language-infusion sub-network) to predict the word-level average BERT embeddings. Only frames corresponding to word end times $tw$ (obtained from force-alignment) are used to minimize the L2 loss between the frame-level predictions and the corresponding word-level embeddings. When using explicit language layers, we do not freeze the HuBERT weights during training. In our experiments, we 
show that we can directly infuse language information into HuBERT weights without using language layers. Let $b_e$ represent the BERT embeddings for tokens in the word, then the word embedding $embed_{w}$ is defined as:
\begin{equation}
embed_{w} = mean(b_e)
\end{equation}
The loss to train the language-infused network is: 
\begin{equation}
L_{utt} = \Sigma_{words}
((p(f_{tw}) - embed_{w})^2)
\end{equation}
where $f_{tw}$ is the closest frame corresponding to the word ending, $p$ is the output of the network. 

When using language layers, the representations of the final language-infused transformer layer and HuBERT's final layer are concatenated and used in place of the HuBERT representations of the downstream task. When no language layers are used, we use the language infused HuBERT representations as inputs for barge-in classification. This technique has similarities to SpeechBERT \cite{chuang2019speechbert}, however SpeechBERT still requires ASR at the inference stage. Our technique is closer to SPLAT \cite{chung2020splat}, but SPLAT only matched the CLS token embedding of BERT with the first frame output of speech representation, which we believe may not lead to as strong an association between the two modalities as in our model. We plan to compare the efficacy of SPLAT and our approach in the future.

\begin{figure}[t] 
  \centering
  \includegraphics[scale=0.65]{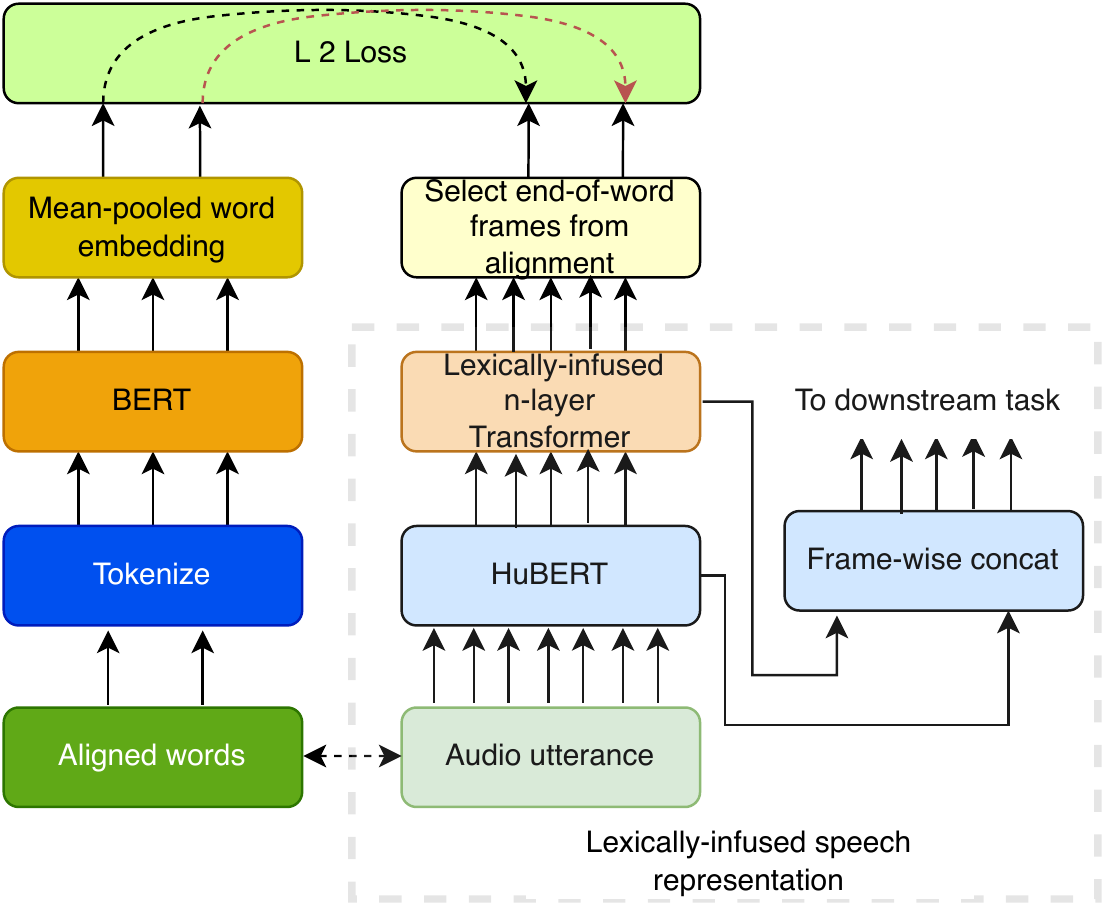}
  \caption{Training to inject lexical information into speech representation}
  \label{fig:model2}
  \vspace{-5mm}
\end{figure}

\section{Experiments}
\label{sec:experiments}

\subsection{Data} 
\label{ssec:data}

Since spoken dialogue systems can be utilized in sensitive public environments, we did not consider home device directed speech data sets for experimentation. Keeping the barge-in task in mind, we have collected 12000 single turn chat-bot prompts, and user response pairs in various environments such as home, car, malls, and streets. The chat-bot prompt consists of multiple simulated questions and users respond to those questions through a simulated false or true barge-in. Users are asked to simulate a true interruption by cutting off the chat-bot with chat-bot directed utterances and false interruptions by speaking random utterances. To create realistic dialog simulation, we provided users with a large list of false interruptions and uncorrelated utterances to choose from and asked them to mimic non chat-bot directed speaking style. For each of these single turn pairs, we also annotate an associated dialogue context. Out of the total 12000 data pairs collected, we made three splits: train (9000), validation (1000) and test (2000). All three splits have balanced true and false barge-in classes and the audio data is upsampled to 16kHz (for experimentation with pretrained models) with each audio spanning an average of 2.4 seconds \cite{guyon1998size}.

\subsection{Model Configurations}
\label{ssec:models}


For baseline systems, we trained a VAD model using TDNN architecture \cite{sugiyama1991review} and a multi-layer LSTM model for barge-in classification. For speech representation, HuBERT \cite{hsu2021hubert} offers three different configurations with varying number of parameters, transformer encoder blocks and embedding size: HuBERT Base (95M, 12 layers, 768), Large (317M, 24 layers, 1024), and X-Large (964M, 48 layers, 1280). Keeping latency in mind, we have chosen the HuBERT Base model which was pretrained on 16kHz Librispeech standard 960hr training data \cite{panayotov2015librispeech} for experimentation. The pre-trained model used for chat-bot prompt representation is a 12-layer \textit{BERT-base-uncased} model with 768 dimensional output from Huggingface \cite{huggingtrans}. For dialogue context representation, we used a trainable embedding layer with 64 dimensional output. We project all three representations to 128 dimensional vectors using separately trained linear layers. For training barge-in models, Stochastic Gradient Descent (SGD) with a learning rate of 5e-4 and a dropout of 0.2 was applied. 
The \textit{Lexically-Infused HuBERT} (LI-HuBERT) model used the same Librispeech pre-trained Hubert-Base model and the BERT model was frozen during pre-training. The transformer layers of the core language-infusion sub-network are similar to that of transformer layers in the HuBERT-Base model. We experimented with and without additional language layers where we infuse language information directly into the HuBERT-Base model. For LI-HuBERT pre-training, we used an in-house conversational dataset consisting of 90k utterances, with each utterance averaging 5s and the word time-stamps for the same are generated from a Kaldi-based hybrid ASR model \cite{Povey_ASRU2011}. For training, we used ADAM optimizer with a learning rate of 2e-4, and gradient norm was clipped to 5. The model was trained for 800k steps, with a batch size of 16. 

To measure latency, we used an AWS EC2 M5 instance with 64GB memory. We utilized the Torch library to load the models on CPU that were previously trained and saved on GPU. We evaluated each model on the aforementioned test split by setting the batch size to 1 and the latency numbers shown in the results section are measured in milliseconds (ms).


\section{Results and Discussion}
\label{sec:results}

The performance of each system is measured by metrics such as average recall and F1 along with latency. First, we present results for a baseline barge-in system with audio (filter-bank features) and ASR hypotheses as inputs in Table \ref{tab:base}. We observe that using a VAD alone doesn't solve the problem as it can not disambiguate the cases of speech not intended for the chat-bot further motivating barge-in classification task. Also, we observe that using ASR hypotheses along with audio provides significant improvement over a pure audio based approach showing that lexical information plays an important role in barge-in classification. However, the latency is increased by 10-fold from 20ms to 226ms as the computation includes the time for ASR decoding and therefore we do not consider ASR hypotheses for rest of the experiments in this paper.

 Table \ref{tab:hubert} provides the results for the barge-in classification task with both frozen and fine-tuned HuBERT representations. As expected, the model trained with HuBERT frozen representations did not perform that well as it was pre-trained only on Librispeech data and is quite diverse from spoken dialogue data used for this task. The fine-tuned HuBERT model using only audio as input outperformed the baseline system that uses lexical information from ASR hypotheses in terms of both F1 (4.5\% relative improvement: 73.6 to 76.9) and latency (reduced from 226ms to 140ms). 


\begin{table}[t]
\caption{Performance of VAD and baseline Barge-in system with audio (filterbank features) and ASR hypotheses as inputs}
\vspace{-3mm}
  \centering
  \small
    \begin{tabular}{p{1.4cm}lccc}
        \toprule
        Model & Input & Recall & F1 & Latency \\
        \midrule
        TDNN (VAD) & Audio & 51.6 & 65.4 & - \\
        \midrule
        LSTM & Audio & 60.7 & 61.3 & 20 \\
        & ASR Hypothesis & 72.1 & 72.5 & 219 \\
        & \hspace{2mm}+ Audio & 73.0 & 73.6 & 226 \\
     \bottomrule
    \end{tabular}
    \label{tab:base}
\vspace{-5mm}
\end{table}

\begin{table}[h]
\caption{Performance of HuBERT based barge-in model augmented with contextual features}
\vspace{-3mm}
  \centering
  \small
    \begin{tabular}{p{1.6cm}lccc}
        \toprule
         Model & Input & Recall & F1 & Latency \\
        \midrule
        HuBERT (Frozen) & Audio & 60.6 & 62.3 & 140 \\
        & + Bot Prompt & 63.5 & 64.0 & 172 \\
        & + Dialogue Ctx  & 65.0 & 64.0 & 143 \\
        & \hspace{2mm} + Bot Prompt  & 64.6 & 65.9 & 176 \\
        \midrule
        HuBERT (Fine-Tuned) & Audio & 75.7 & 76.9 & 140 \\
        &  + Bot Prompt & 76.1 & 76.9 & 172 \\
        &  + Dialogue Ctx  & 76.6 & 75.0 & 143 \\
        &  \hspace{2mm} + Bot prompt & 77.6 & 77.7 & 176 \\
     \bottomrule
    \end{tabular}
    \label{tab:hubert}
    \vspace{-3mm}
\end{table}

\begin{table}[t]
\caption{Performance of Lexically Infused HuBERT (LI-HuBERT) with language layers and contextual features}
\vspace{-3mm}
  \centering
  \small
    \begin{tabular}{p{2cm}lccc}
        \toprule
         Model & Input & Recall & F1 & Latency \\
         \midrule
        LI-HuBERT (Fine-Tuned) & Audio & 78.8 & 79.1 & 140 \\
        + 2 layers &  & 79.7 & 79.9 & 144 \\
        + 4 layers  &  & 80.2 & 80.3 & 152 \\ \midrule
        LI-HuBERT (Fine-Tuned) & + Bot Prompt & 80.2 & 81.1 & 172 \\
        & + Dialogue Ctx  & 78.9 & 79.0 & 143 \\
        & \hspace{2mm} + Bot Prompt   & 81.3 & 81.3 & 176 \\
     \bottomrule
    \end{tabular}
    \label{tab:dial}
    \vspace{-5mm}
\end{table}

%
%

\noindent \textbf{Performance with contextual information:} We now compare the performance of our HuBERT based barge-in models when augmented with spoken dialogue contextual features. From Table \ref{tab:hubert}, we observe that adding contextual cues such as bot prompt and dialogue context helps improve the performance of both frozen and fine-tuned HuBERT models. With fine tuning, the best performing model with both dialogue context and bot prompts achieves 2.5\% relative improvement in recall (improved from 75.7 to 77.6). This shows that the bot prompt and dialogue contexts aid in classification, especially when they have correlation with the user responses. Though adding these contextual representations led to an increase in latency, the best performing model is still 22\% faster than the baseline model that uses ASR hypotheses.

\vspace{2mm}
\noindent \textbf{Performance with language infusion:} In this section, we discuss the improvements obtained from pre-training with language infusion. From Table \ref{tab:dial}, we observe that our proposed \textit{LI-HuBERT} model to validate barge-ins directly from the audio is not only significantly better (7.5\% relative F1 improvement: 73.6 to 79.1) but also 38\% faster than a baseline system that uses lexical information from ASR hypotheses. We also observe that adding additional language layers for language infusion yields improvements in barge-in classification with increased latency. We can attribute these improvements to the additional parameters incurred with additional layers to learn language information. Finally, we combine the contextual features with our \textit{LI-HuBERT} model in order to validate if each of the techniques would yield additional gains when applied together. From Table \ref{tab:dial}, it is clear that using lexically infused speech representations with contextual embeddings provides us with the best performance (10.4\% relative F1 improvement: 73.6 to 81.3). Overall, the improved performance suggests that language infusion can help carry sufficient semantic information to distinguish between true and false barge-ins. 






\noindent \textbf{Qualitative Analysis:}  For further analysis, we listened to some of the audios from test data pertaining to misclassified examples from each model. We observed that the VAD model has very high false barge-in rate (indicated by low average recall), as it cannot differentiate non chat-bot directed utterances. The baseline model that uses ASR hypotheses has relatively poor true barge-in detection rate, especially in cases of longer utterances due to ASR errors. 
Our proposed model has better detection under these conditions as the model does not rely on ASR performance, and rather relies on pretraining step for linguistic knowledge infusion. 
However, our proposed model has poor false barge-in classification performance under certain noise conditions (e.g., audio cuts) and can be mitigated to a certain degree by performing data augmentation with external noise during training, and we will pursue this as future work.


\section{Conclusion}
\label{sec:conclusion}
In this work, we introduced the task of contextual acoustic barge-in where we detect user interruptions in close-talk environments. For this task we have proposed a classification approach using HuBERT features. Our experiments have shown that fine-tuned HuBERT features achieve better performance compared to an ASR hypothesis based models with significantly lower latency. We have also shown that using dialogue information available from a chat-bot prompt and dialogue context further improves performance with some latency hit. We further proposed a new technique for infusing language information over HuBERT representations which gave us the best performance. In our future work, we will explore further improving the robustness of existing solution.


\bibliographystyle{IEEEtran}

\bibliography{mybib}

\end{document}